%% file: bare_jrnl.tex
\newcommand{\cmark}{\ding{51}}%
\newcommand{\xmark}{\ding{55}}%

\newcommand{\miaojing}[1]{{#1}}

\documentclass[journal]{IEEEtran}
\ifCLASSINFOpdf
   \usepackage[pdftex]{graphicx}
\else
   \usepackage[dvips]{graphicx}
\fi

\usepackage{graphicx}
\usepackage{amsmath}
\usepackage{amssymb}
\usepackage{subfigure}
\usepackage{multirow}
\usepackage{tabularx}
\usepackage{booktabs}
\usepackage{diagbox}
\usepackage{paralist}
\usepackage{epstopdf}
\usepackage{color}
\usepackage{pifont}

\input{macro}
\begin{document}
%
\title{Dam reservoir extraction using point-level and prior-guided metric learning}
\title{Dam reservoir extraction from high-resolution remote sensing imagery using tailored metric learning strategies}
\title{Dam reservoir extraction from remote sensing imagery using tailored metric learning strategies}
%
%
%

\author{
        Arnout van Soesbergen$^*$,~
        Zedong Chu$^*$,~
        Miaojing Shi$^\dagger$,~\IEEEmembership{Member,~IEEE,}
        and~Mark~Mulligan~
\thanks{$^*$Equal contribution with alphabetic order. $^\dagger$Corresponding author.}
\thanks{Arnout van Soesbergen and Mark Mulligan are with the Department of Geography, King's College London. Arnout van Soesbergen is also with the UN Environment World Conservation Monitoring Centre, Cambridge, UK Email: \{arnout.van\_soesbergen, mark.mulligan\}@kcl.ac.uk }
\thanks{Zedong Chu and Miaojing Shi are with the Department of Informatics, King's College London. E-mail: chuzedong98@gmail.com, miaojing.shi@kcl.ac.uk}
}


%
%

\markboth{Accepted on IEEE Transactions on Geoscience and Remote Sensing}%
{Soesbergen \MakeLowercase{\textit{et al.}}}
%



\maketitle

\begin{abstract}
Dam reservoirs  play an  important  role  in  meeting  sustainable  development goals and global climate targets. However, particularly for small dam reservoirs, there is a lack of consistent data on their geographical location. 
To address this data gap, a promising approach is to perform automated dam reservoir extraction based on globally available remote sensing imagery. It can  be considered as a fine-grained task of water body extraction, which involves extracting water areas in images and then separating dam reservoirs from natural water bodies.
A straightforward solution is to extend the commonly used binary-class segmentation in water body extraction to multi-class. This, however, does not work well as there exists not much pixel-level difference of water areas between dam reservoirs and natural water bodies.
We propose a novel deep neural network (DNN) based pipeline that decomposes dam reservoir extraction into water body segmentation and dam reservoir recognition. {Water bodies are firstly separated from background lands in a segmentation model and each individual water body is then predicted as either dam reservoir or natural water body in a classification model.} For the former step, point-level metric learning with triplets across images is injected into the segmentation model to address contour ambiguities between water areas and land regions. For the latter step, prior-guided metric learning with triplets from clusters is injected into the classification model to optimize the image embedding space in a fine-grained level based on reservoir clusters.
To facilitate future research, we establish a benchmark dataset with earth imagery data and human labelled reservoirs from river basins in  West  Africa  and  India. Extensive experiments were conducted on this benchmark in the water body segmentation task, dam reservoir recognition task, and the joint dam reservoir extraction task. 
Superior performance has been observed in the respective tasks when comparing our method with state of the art approaches.
 The codes and  datasets are available at \textcolor{magenta}{ \emph{https://github.com/c8241998/Dam-Reservoir-Extraction}}.
\end {abstract}

\begin{IEEEkeywords}
Dam reservoirs, remote sensing, water bodies, segmentation, classification, metric learning.
\end{IEEEkeywords}

%
\IEEEpeerreviewmaketitle

\section{Introduction}\label{Sec:intro}
%
%
%
%
\IEEEPARstart{D}{am} reservoirs are water bodies bounded by artificial dams. {M}{illions} of them exist worldwide, ranging from small farm ponds to very large multi-year storage dams built in large rivers. Many more are being planned or under construction~\cite{mulligan2020goodd,zarfl2015global}. These structures store water for irrigation and domestic uses, hydropower generation and flood protection. As such, dam reservoirs play an important role in meeting the sustainable development goals (SDGs) and global climate targets. On the other hand, they can also have negative impacts on social and ecological systems, e.g. river fragmentation~\cite{grill2019mapping}.

\begin{figure}[t]
\centering
\includegraphics[scale=.38]{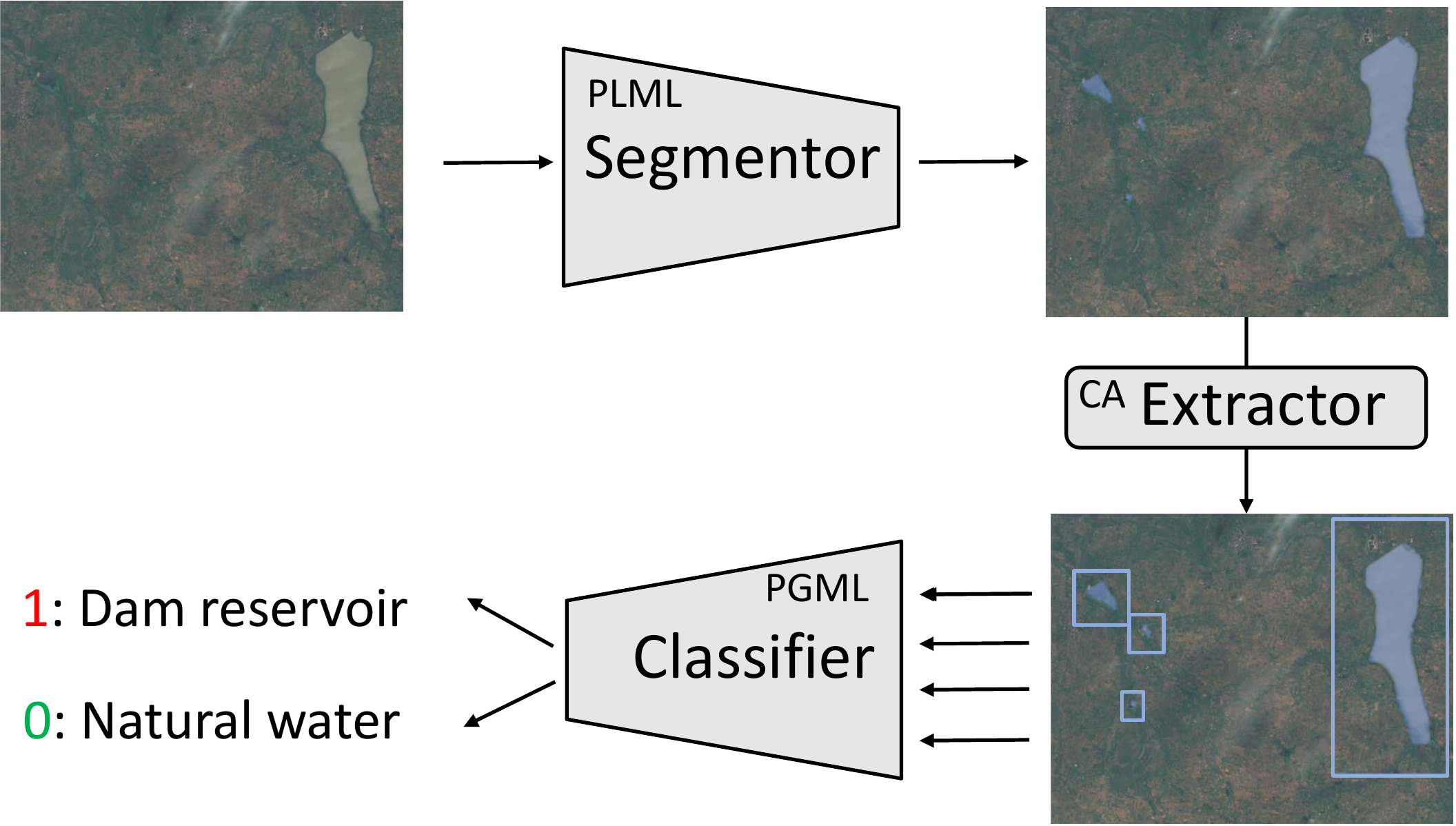}
\caption{Overview of the proposed dam reservoir extraction pipeline. Given an input image, a water body segmentor is firstly introduced to segment water bodies from the image. Afterwards, a connected area (CA) extractor is devised to extract individual water bodies with bounding boxes. Finally, each bounding box image is sent into a water body classifier to get its label as either dam reservoir or natural water body. PLML and PGML refer to point-level and prior-guided metric learning, respectively. }
\label{fig:wholepipeline}
\vspace{-2mm}
\end{figure}

Despite the recognized importance and social and environmental concerns with dams, there is still a lack of consistent data on their geographical distribution, particularly for smaller dams~\cite{de2019beware,belletti2020more}. Some of the largest publicly available datasets only include around 38,000 dams with precise spatial locations, yet it is estimated that there may be millions of dams worldwide~\cite{lehner2011high}.
This lack of data limits our understanding of impacts of dams and thus our ability to monitor and manage progress towards sustainability, climate and biodiversity goals. To address this challenge, as recently noted by the Global Dam Watch initiative~\cite{mulligan2021global}, the most promising way is to perform automated dam reservoir extraction based on globally available remote sensing imagery and novel computing techniques.  
It allows for the extraction of smaller dam reservoirs and provides a mechanism to regularly and consistently update datasets with newly built or removed dam reservoirs. 

{Dam reservoir extraction aims to extract surface extents of both dams and associated water areas (\eg ponds, rivers) in satellite imagery.}
It can be considered as a fine-grained task of water body extraction, where we need to further separate dam reservoirs from  natural water bodies. 
Building upon the development of deep neural networks (DNN)~\cite{krizhevsky2012imagenet}, many methods solve water body extraction as a binary-class segmentation task~\cite{zhang2021tgrs,feng2018grsl,miao2018grsl}. Intuitively, our work can be solved as a multi-class segmentation task among dam reservoir, natural water body, and background land. This, however, does not work well in practice as there exists not much pixel-level difference of water areas between dam reservoirs and natural water bodies (see experiments in Sec.~\ref{sec:exp-overall}).
One major difference between dam reservoirs and natural water bodies lies in their shapes, where dam reservoirs tend to have distinct shapes along the dam side (\eg straight dam walls) whilst natural water bodies usually do not.
In this paper, we describe a novel approach that decomposes dam reservoir extraction into two tasks: water body segmentation and dam reservoir recognition; both are binary missions. For the former, we employ the DeepLabV3+~\cite{chen2018eccv} to segment water areas against background lands. For the latter, we employ the ResNet50V2~\cite{he2016identity} to classify each segmented water area as either dam reservoir or natural water body. 

In practice, contour ambiguities between water areas and land regions impede the segmentation while color similarities between natural water and dam reservoirs complicate the classification. To tackle these issues, the idea of metric learning is exploited in both tasks with two distinct contributions: for water body segmentation, we propose to implement the metric learning using the point-level feature representations of the model. A triplet loss is formulated with hard positive/negative pairs chosen from mis-predicted points. Specifically, we allow to construct the triplet across images to improve the robustness of the segmentation against complex backgrounds and seasonal changes. The proposed point-level metric learning (PLML) module enforces the similarity of points from water areas and the dissimilarity of them between water areas and land regions. For dam reservoir recognition, we choose the image-level metric learning with triplet loss rather than the classic cross entropy loss as the former performs better in addressing the difference between dam reservoirs and natural water bodies. We observe that many water bodies tend to fall within a number of subgroups according to their distinct attributes, for instance, the shape of water bodies, which can be triangular, rounded, linear, or irregular.
This shape information is implicitly encoded into the image feature embedding in the convolutional neural network (CNN). To optimize the embedding space in a fine-grained level, we introduce a prior-guided metric learning (PGML) module: features of training samples are clustered into several groups on the fly such that those of the same class and cluster are pushed close while those of different classes are pulled away in a triplet loss.

For a complete pipeline, as shown in Fig.~\ref{fig:wholepipeline}, the water body segmentation results obtained from the first task are fed as inputs to the second task to classify as either dam reservoirs or natural water bodies. To the best of our knowledge, this work is the first that focuses on dam reservoir extraction using deep learning techniques. To facilitate future research in this field, we create a benchmark dataset from the earth imagery data of the Volta river basin in West Africa and the Cauvery river basin in India with labelled dam reservoirs available.
The two tasks are conducted either independently or jointly in this dataset. Sizeable benefits are observed by comparing our method to state of the art  in the respective task.

\begin{figure*}[htb]
\centering
\includegraphics[scale=.43]{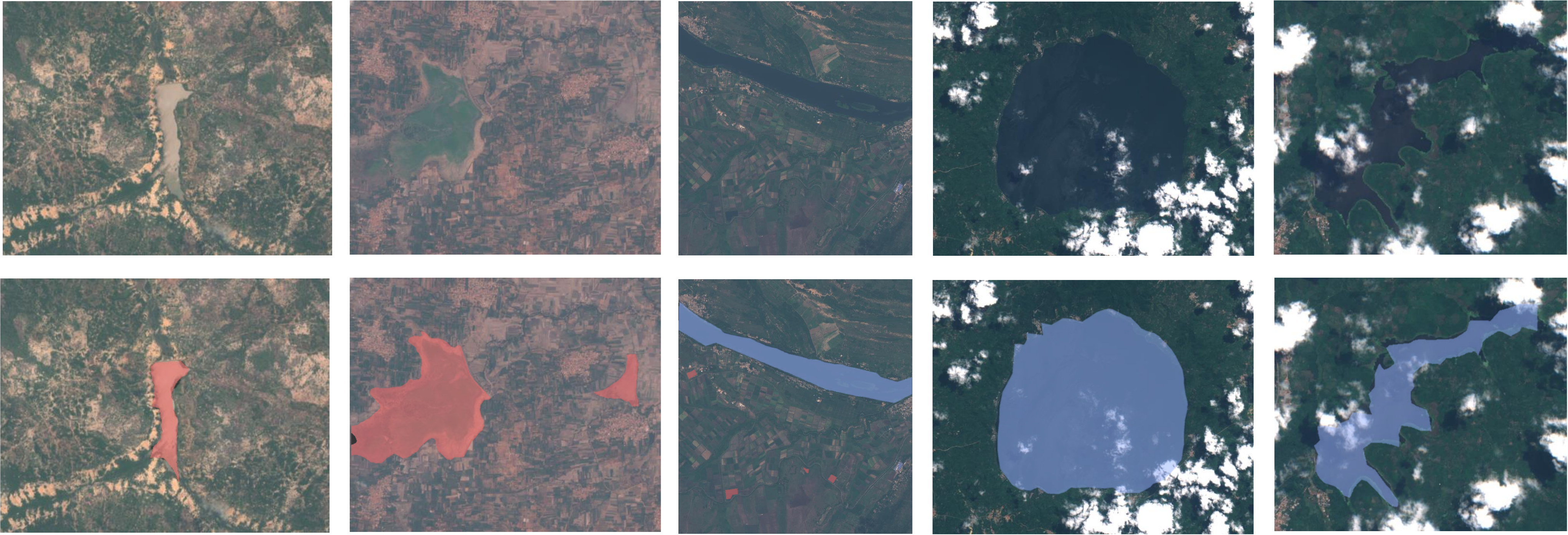}
\caption{Samples from Sentinel imagery of the  Volta  river  basin  in  West  Africa (column 1,3,4,5) and  the  Cauvery  river  basin  in  India (column 2). The red and blue masks are the ground truth labels {(digitised manually from high resolution imagery)}  indicating the dam reservoir and the natural water, respectively. 
}
\vspace{-2mm}
\label{fig:data}
\end{figure*}

\section{Related works}

Water body extraction is essential to environmental monitoring and has been widely explored in the literature of computer vision and remote sensing~\cite{ding2014grss,Liu2016GRSL,maggiori2016tgrs,feng2018grsl,zhang2021tgrs}. %
Its challenge lies in both the complex background of land regions and lower color intensities of water areas~\cite{zhang2021tgrs}. Early methods tackle this problem by applying unsupervised techniques \ie thresholding~\cite{ranjani2012rsni,amitrano2014ijst}
clustering~\cite{Liu2016GRSL} 
or edge extraction~\cite{felix2014rs} \etc
These methods perform well at small scales yet fail to generalize on water bodies of many variations.
With the advent of deep learning, modern methods solve water body extraction as a supervised binary segmentation task~\cite{maggiori2016tgrs,miao2018grsl,feng2018grsl,li2021deep,heidler2021tgrs,zhang2021tgrs}. %
Based on the fully convolutional network (FCN)~\cite{long2015tpami}, Maggiori ~\etal~\cite{maggiori2016tgrs} first designed an end-to-end framework to provide accurate pixel-wise classification for satellite images. Miao~\etal~\cite{miao2018grsl} proposed a restricted receptive field deconvolution and an edges weighting loss (EWLoss) for water body segmentation in remote sensing imagery. Feng \etal~\cite{feng2018grsl} presented a deep U-Net architecture~\cite{Ronneberger2015miccai} by applying super-pixel segmentation and conditional random fields (CRFs) for water body extraction.
Recently, 
Zhang~\etal~\cite{zhang2021tgrs} integrated the fully convolutional upsampling pyramid network with fully convolutional conditional random fields to realize water body extraction in high-resolution SAR images.
Metric learning has been employed in remote sensing but is mostly on the image-level~\cite{zhang2021ijst,kang2020tgrs}.
To improve the robustness of our segmentation model,
we introduce a point-level metric learning scheme which constructs triplets from correctly predicted water points and mis-predicted points across images. \miaojing{Besides deep learning-based methods, there are also good remote sensing only approaches for water body extraction, \eg through inclusion of additional spectral bands (see ~\cite{bijeesh2020surface} for a review). However our proposed pipeline has wider potential for transferability as it can more easily be applied to other imagery sources from satellite, unmanned aerial vehicle (UAV) or historical imagery for which additional bands may not be available or only at lower resolution. In addition, the proposed pipeline provides an end-to-end learning capability that reduces pre-processing efforts on the imagery source.}


There are also many studies that have focused on image classification in  remote sensing {using deep learning techniques}~\cite{Marmanis2016grsl,chen2016tgrs,cheng2018tgrs,gong2017tgrs,zhang2019tgrs,wang2020tgrs,chen2021tgrs,fang2019recognizing}.  %
Cheng~\etal~\cite{cheng2018tgrs} imposed a metric learning regularization on the CNN features of images to make them more discriminative for classification. Wang~\etal~\cite{wang2020tgrs} presented an architecture search framework to automatically find the most appropriate CNN for image recognition in high spatial resolution remote sensing scenes. Similarly, Chen~\etal~\cite{chen2021tgrs} proposed an architecture learning framework combined with channel compression to learn the task-oriented CNN architecture.
Gong~\etal~\cite{gong2017tgrs} utilized the structural information within training batches as the diversity-promoting prior to diversify learned parameter factors for  metric learning in remote sensing image classification.

\miaojing{For dam reservoir-related tasks, there are few works in the literature. Balaniuk~\etal~\cite{balaniuk2020mining} employed a FCN for the segmentation of tailings dams against background lands, which is a binary segmentation task.
Fang~\etal~\cite{fang2019recognizing} employed a ResNet50 network to recognize global reservoirs in remote sensing imagery. Malerba~\etal~\cite{malerba2021continental} utilized ResNet34 to perform farm dam classification upon satellite images of Australian territories. ~\cite{fang2019recognizing,malerba2021continental} are basically a binary classification task. Compared to~\cite{balaniuk2020mining,fang2019recognizing,malerba2021continental}, ours is a fine-grained and more challenging task that we first perform a binary segmentation between water body and background land and then carry out a binary classification between natural water body and dam reservoir.
Our architecture is also more complicated: we build on two base networks, DeepLabV3+ for segmentation and ResNet50V2 for classification. We introduce a novel point-level metric learning strategy to improve DeepLabV3+ and a novel prior-guided metric learning strategy to improve ResNet50V2. }

\begin{figure*}[t]
\centering
\includegraphics[scale=.3]{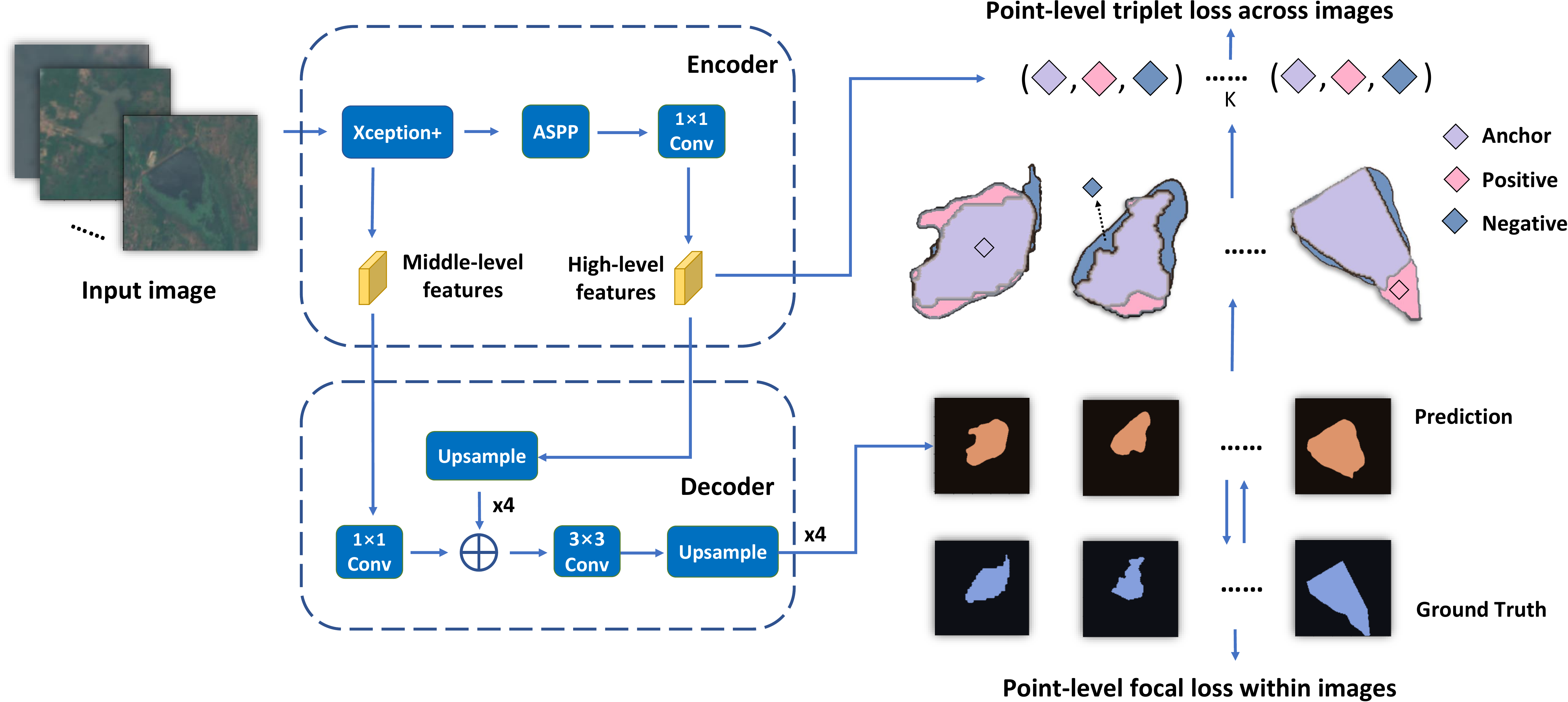}
\caption{Architecture of the proposed segmentation model.
The backbone is DeepLabV3+ (left side). A point-level metric learning scheme with triplets across images is devised upon it (right side). During training, based on the comparison between the prediction and ground truth,  correctly predicted pixels, false negative and false positive pixels are extracted from each image. These three sets of pixels are merged across images in a training batch, respectively. Triplets are constructed by randomly selecting the anchor, hard positive, and hard negative points from the three pools correspondingly. The model is optimized by the point-level triplet loss across images and the point-level focal loss within images at the same time. For more details please refer to Sec.~\ref{sec:water_body_segmentation}.
}
\label{fig:pipeline}
\end{figure*}

\section{Data Source}\label{sec:watersource}

\subsection{Study area}\label{sec:data-area}
Our study focuses on the Volta river basin in West Africa and the Cauvery river basin in India for model training and testing. The Volta basin is a large (~413,000 sq km)  trans-boundary river basin, draining parts of six countries in West Africa and is characterized by unevenly distributed rainfall along a north-south gradient (400mm/yr in the north to over 1700mm/yr in the south). The basin contains a large number of small reservoirs (88\% of reservoirs are smaller than 1 Mm$^3$~\cite{jones2017big}) which are challenging to map using traditional remote sensing approaches. The Cauvery River basin in southern India is a monsoonal basin draining an area of 81,155 km$^2$. It has a west to east decreasing precipitation gradient, influenced by the SW monsoon winds. Precipitation varies between 300 mm/yr in the east of Karnataka plateau to nearly 6,000 mm/yr in the Western Ghats~\cite{meunier2015controls}. The basin contains many small reservoirs used for irrigation, estimated to provide irrigation water to some 2 million hectares~\cite{stacey2013water}. Within each river basin, we extracted rectangular samples of varying sizes in a GIS (QGIS 3.6). \miaojing{Samples of varying sizes enable a diversity of water bodies in them, which benefits the model's robustness and generalizability. Also, we aim to find multiple dam reservoirs in each sample instead of finding just one. As a result, it is necessary to adjust the sample size to include several water bodies. Lastly, for some big water bodies, a sample of fixed size can only include a partial of them, making the water bodies incomplete.} In the end, a total of 193 samples were created from the Volta basin and 102 from the Cauvery basin.


\subsection{Remote sensing data}
We used Sentinel 2, \miaojing{ Level-1C} data  at 10 metre resolution as this is the highest resolution freely available for remote sensing imagery. We utilised Google Earth Engine \miaojing{\cite{gorelick2017google}} through the Earth Engine Python API to create cloud free mosaics, removing images with more than 5\% cloud coverage using the QA 60 cloud mask band. Mosaics were created for each data sample \miaojing {for the period 2017-2020} and returned as RGB images, utilizing the B4, B3 and B2 bands. For the Cauvery river basin in India, many images contain clouds. For the Volta river basin in West Africa, only a small number of images contain clouds. 
Some examples are shown in Fig.~\ref{fig:data}.

To aid the identification of natural and dam water bodies, we used the Joint Research Centre (JRC) global water body dataset V1.3 ~\cite{pekel2016high} at 30 metre resolution also accessible through Google Earth Engine. We used the max extent product which provides a binary image containing 1 anywhere water has ever been detected based on Landsat 5, 7 and 8 data for the period 1984-2020. These data were obtained for the full extent of the river basins.

\subsection{Labels}
Pixel labels for natural and dam reservoir outlines were obtained from OpenStreetMap (OSM) using the overpass API\footnote{https://overpass-turbo.eu/} from the QGIS 3.6 QuickOSM plugin for the full extent of both river basins.
These labels were manually verified and corrected where necessary. Using the JRC global water body dataset~\cite{pekel2016high}, associated water areas on existing dams were localised  from~\cite{jones2017big} for the Volta basin and from GOODD~\cite{mulligan2020goodd}, GRanD~\cite{lehner2011high} and OpenStreetMap\footnote{www.openstreetmap.org} for the Cauvery basin. Missing reservoirs and lakes were manually digitised for all samples. Additional dam features for all dams in these basins are available in the Global Dam Watch Knowledgebase\footnote{ www.globaldamwatch.org}.
Within the 193 samples in the Volta basin, 217 reservoirs and 99 natural water bodies were identified. Within the 102 samples in the Cauvery basin, 351 reservoirs and only 3 natural water bodies were identified. There are very few natural water bodies in the Cauvery basin, we therefore merged samples from both Volta and Cauvery to relieve this data imbalance.
The natural and dam reservoir labels (masks) were rasterized and converted to TIF format with a resolution compatible with the Sentinel-2 image data. 
As shown in Fig.~\ref{fig:data}, natural reservoir areas are labeled as 1 (blue) and dam reservoir areas are labeled as 2 (red).

\section{Methodology}

This study aims to extract water bodies in remote sensing imagery and classify them as either natural water bodies or dam reservoirs. We first introduce the water body segmentation task and then the dam reservoir recognition task. In the end, we present a pipeline that uses a connected area extractor to connect both tasks as one joint mission.

\subsection{Water body segmentation}\label{sec:water_body_segmentation}

\para{Base model.}
We use DeepLabV3+~\cite{chen2018eccv} as our base model for the binary segmentation of water bodies. It has an encoder-decoder structure where the encoder projects the image into a latent feature while the decoder predicts the segmentation result. The encoder starts with an improved Xception structure (Xception+), which is based on the depth-wise separable convolution. In order to capture the contextual information over multiple scales, the atrous spatial pyramid pooling (ASPP) module is implemented. It contains five layers: three 3$\times$3 atrous conv layers, one 1$\times$1 conv layer and one global average pooling layer. The atrous convolution adjusts the size of filters to extract multi-scale information from the encoder. Features over different scales are concatenated into a new 256-dimensional feature. It passes through another 1$\times$1 conv layer to generate the final feature representation (the high-level feature in Fig.~\ref{fig:pipeline}).
The decoder concatenates both high-level and middle-level features where the latter is taken from the end of the Xception+. The high-level feature is upsampled to the same size of the middle-level feature while the channel of the middle-level feature is reduced by an 1$\times$1 conv to not outweigh the high-level feature. Their fused feature is passed through a 3$\times$3 conv layer and upsampled to the original size of the input image to produce the final segmentation result.


\para{Point-level metric learning with triplets across images.} ~The base model faces challenges such as contour ambiguities between water areas and land regions.  
A particular reason causing this is seasonal changes on water areas. For instance, the visible water area in an image taken in the dry season will be smaller than in the rainy period, which leads to the ambiguity between water and land on the ground truth boundaries. Fig.~\ref{fig:data}: {second} column provides an example of this. To deal with this, we introduce a point-level metric learning (PLML) with triplets across images. It specifically optimizes the model on ambiguous areas/points between water and land and therefore improves the robustness to seasonal changes.


We implement the metric learning module in the form of a triplet network. Triplets are constructed as follows: during training, given an image prediction and its ground truth, water points that are correctly predicted are considered as a pool of anchors; water points that are mis-predicted as land and land points that are mis-predicted as water are considered as pools of hard positives and hard negatives, respectively. 
The latter two pools contain mostly ambiguous points for water and land. We can correspondingly select an anchor $a$, a hard positive $p$ and a hard negative $n$ from the three pools. The triplet loss pushes $a$ and $p$ close while $a$ and $n$ away:
\begin{equation}\label{eq:seg_triplet}
    L_\text{triplet}(a,p,n)=\max (d(a, p) - d(a, n) + \beta, 0),
\end{equation}
we take the high-level feature of each point from the backbone (see Fig.\ref{fig:pipeline}) and compute their Euclidean distance in $d(\cdot,\cdot)$. We request $d(a, p)$ to be smaller than $d(a, n)$ with a margin of $\beta$.

\begin{figure*}[htb]
\centering
\includegraphics[scale=.42]{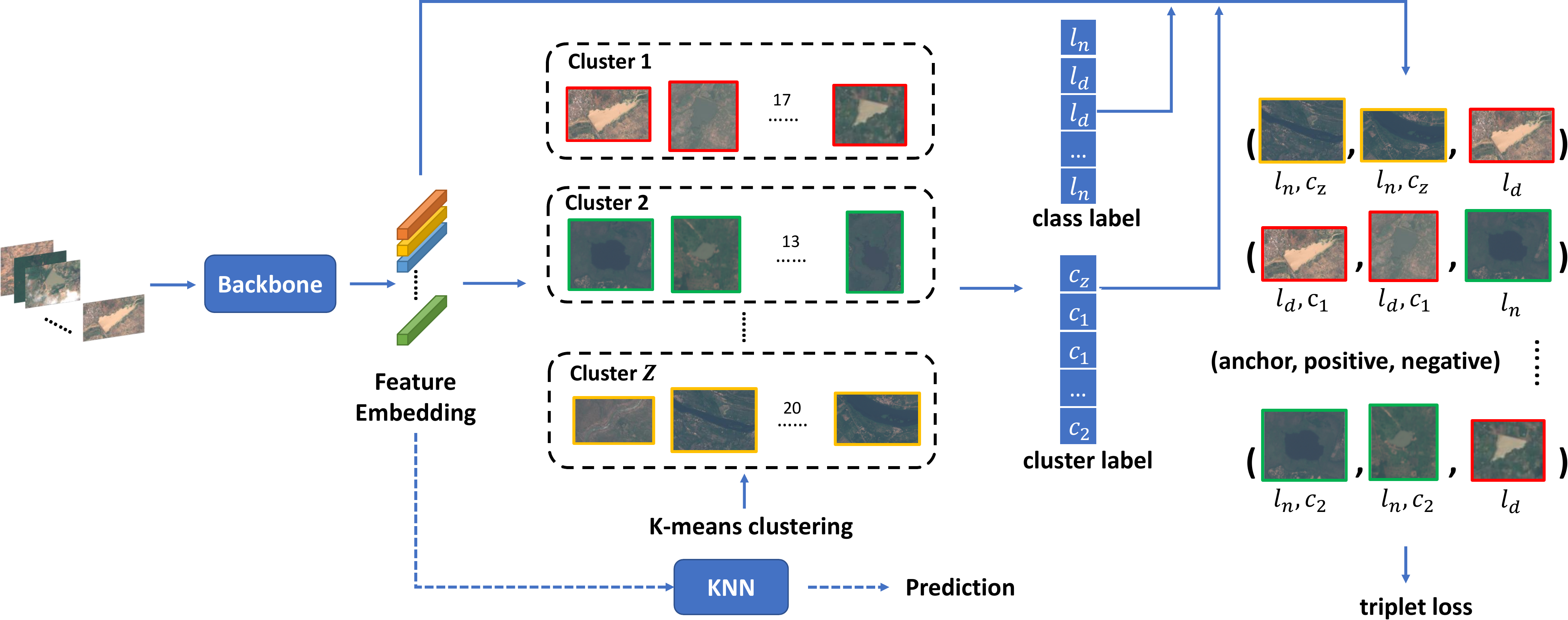}
\caption{Architecture of the proposed classification model. The backbone is ResNet50V2. A prior-guided metric learning with triplets from clusters is devised upon it. Within each batch, images are first embedded in the backbone to obtain global features. A K-means clustering is applied on the fly on these features to separate them into different groups. According to the cluster label and the class label, triplets can be constructed to push images of the same class and cluster close and images of different classes away in the feature space. }
\label{classification}
\end{figure*}

Instead of constructing triplets within one image, we allow this construction across images. The anchors, hard positives, and hard negatives are merged respectively over a set of images in a training batch.
Given one anchor point, it has the opportunity to be paired with hard positives/negatives from other images of different territories or seasonal appearances, therefore improving the model's generalizability over different testing scenarios. Formally, over a batch of $B$ images, we first randomly choose $K$ anchor points from each image. For every anchor, its paired hard positive and negative are chosen from corresponding pools over the entire batch. They can be chosen based on their feature distances to the anchor or simply by randomness. We argue that the hard positive/negative pools are already attached to the false negative/positive predictions of the model, a random selection from corresponding pools is sufficient and efficient.
The total loss in a batch is given by:
\begin{equation}\label{eq:triplet}
{L_{\mathrm{M}}}=\frac{1}{B}\frac{1}{K}\sum_{b=1}^{B}\sum_{k=1}^{K} L^{bk}_\text{triplet}
\end{equation}
within the $k$-th triplet in the $b$-th image, $p$, $n$ can be from different images to that of $a$.  \miaojing{Notice there may be less than $K$ anchors in some images, we choose the actual number of anchors to construct the triplets. Moreover, if images in a batch have no anchors, the triplet loss for this batch returns a value of zero; if an anchor has no hard positives or hard negatives, its triplet will be ignored in the loss function. In summary, we can choose the maximum of $K$ anchors or the minimum of zero anchor from each image. The former is the majority.}

\para{Network loss.}
We use focal loss~\cite{lin2017focal} for the segmentation prediction. For each pixel, the loss is defined as:
\begin{equation}\label{eq:seg_loss}
    L_{\text {focal}}(p, y)=\left\{\begin{array}{cll}
-\alpha(1-p)^{\gamma} \log (p), & \text { if } \quad y=1 \\
-(1-\alpha) p^{\gamma} \log (1-p), & \text { if } \quad y=0
\end{array}\right.
\end{equation}
\miaojing{where $y\in \{0,1\}$ specifies the ground truth class label with 1 for the water class and 0 for the land class. $p \in [0,1]$ is the predicted probability of a sample for the water class. $\alpha$ is a hyper parameter to address the class imbalance problem. $(1-p)^{\gamma}$ is a modulating factor to control the contrast of loss values between easy-predicted pixels and hard-predicted pixels. With a proper ${\gamma}$, it down-weights easy-predicted pixels and up-weights hard-predicted ones. We give an example: assuming that $\gamma$ is 2, if one water pixel has a high predicted probability ($p = 0.9$) for being water, then its modulating factor $(1-p)^\gamma$ is 0.01. This pixel has a high confidence in the prediction, which is regarded as an easy-predicted pixel, its loss value will be reduced by 100 times. However, if one water pixel has a low predicted probability ($p = 0.1$), its modulating factor $(1-p)^\gamma$ becomes 0.81. This is a hard predicted pixel, its loss value will only shrink by a factor of 0.81, which ends up with a bigger weight in the loss function than the former easy predicted pixel.}

The segmentation loss over a batch is:
\begin{equation}
{L_\text{S}}=\frac{1}{B}\frac{1}{N}\sum_{b=1}^{B} \sum_{n=1}^{N} L^{bn}_{\text{focal}}.
\end{equation}
We assume each image has $N$ pixels.
Together with the triplet loss in (\ref{eq:triplet}), we have the total loss:
\begin{equation}\label{eq:loss_seg}
  {L} = {L_\text{S}} + \sigma\times {L_\text{M}},
\end{equation}
$\sigma$ is the loss weight between two terms.

\subsection{Dam reservoir recognition}\label{sec:dam-recog}

\para{Base model.}
We use ResNet50V2~\cite{he2016identity} as our base model for the dam reservoir recognition. ResNet has been proposed to solve the saturation and degradation of accuracy in deep CNN training. Its structure has different groups of res-blocks to learn the residual of the input and output of each network layer. ResNet50V2 has an improved res-block which uses batch normalization before every weight layer and therefore reduces overfitting.

\para{Prior-guided metric learning with triplets from clusters.}
Instead of using the classic cross entropy loss for network optimization, we found that metric learning is empirically more suitable to address variations among natural and dam reservoirs. We remove the last softmax layer in ResNet50V2 but use the 2048-dimensional feature embedding. \miaojing{The network can be optimized with the image-level triplet loss~\cite{schroff2015facenet}: the feature of one image would be pushed close to features of those images of the same class, while pulled away from features of those images of different classes. This essentially works for a basic classification problem but is not sufficient in the dam reservoir recognition. As shown in Fig.~\ref{fig:data}, samples within the same class (\ie dam reservoir or natural water body) often have a big variation (\eg  in terms of shape). They do not necessarily need to be pushed very close together in the feature space despite having the same class. }

\miaojing{We observe that many water bodies tend to fall within a number of subgroups according to their distinct attributes. For instance, one apparent attribute is the shape of water bodies, which can be triangular, rounded, linear, or irregular.}
A triangular water body is likely to be a dam reservoir (\eg a river dammed at one end), a rounded water body is usually a natural lake (\ie natural water body), while linear water bodies tend to be rivers. Of course there are exceptions and much complexity depending on geomorphological and other conditions. Intuitively, water bodies with the same class and similar shapes should be pushed close in the feature space of the network. On the other hand, those with the same class but different shapes might not necessarily need to be pushed close. \miaojing{This observation motivates us to have a better triplet steering in the metric learning. Since shape is only one attribute of water bodies, despite being very important, using it alone is not sufficient to classify between natural water body and dam reservoir. Instead, we propose to cluster samples of water bodies based on their feature embedding in the base model, which encodes not only the shape information but also other cues such as color, texture, appearance, \etc Water bodies of the same class and same cluster are deemed to share many similar attributes and should therefore be pushed close in the feature space.  Based on this, we perform an on the fly feature clustering within each training batch and introduce a prior-guided metric learning (PGML) scheme with triplets from clusters. }




\miaojing{Specifically, given an input image, we first extract its feature in the base model. Features extracted at this stage primarily encode high-level information of objects, \eg shape, size and appearance, but also contain some low-level information, \eg color, texture. To automatically separate images of water bodies into different groups, a K-means clustering is adopted. Because of the large variations of water bodies, we propose to perform the clustering on the fly over image features in a training batch. The clustering becomes a dynamic process adapted to each batch, which can be more accurate. Fig.~\ref{classification} presents the clustering result for a training batch. One can clearly see the visual similarity of samples within each cluster. }
The triplet is constructed by iterating each image in the batch as an anchor. Let $I^b$ be $b$-th image in the batch, as illustrated in Fig.~\ref{classification}, its positive pair should be chosen from images that have the same cluster label ($c_z, z =1,...,Z$) and classification label ({$l_n$} or $l_d$) . Its negative pair should be chosen from images with different classification labels. Specifically, we choose the hard positive image and hard negative image as ones that have the largest/smallest feature distance to the anchor, respectively.

\para{Network loss.}
The triplet loss is utilized to optimize the image feature embedding in the network:

\begin{equation}\label{eq:cls_triplet}
L_\text{C} =  \frac{1}{B} \sum_{b=1}^{B} \max (d(f^b, f^p)-d(f^b, f^n)+\epsilon, 0)
\end{equation}
where $d(\cdot,\cdot)$ computes the distance of image features. $f^b$, $f^p$, and $f^n$ are the features of the anchor, hard positive, and hard negative image, respectively. $\epsilon$ is a margin similar to $\beta$ in (\ref{eq:seg_triplet}). $B$ is the total number of images per batch.
In exceptional cases when an anchor has no candidate positives or negatives, its triplet loss will be ignored. 


\miaojing{At the inference stage, we should have all the feature embedding of training images available through the trained network. Given a new image, it is firstly fed into the network to obtain the feature, then we compute the feature-based cosine similarity between this new image to all training images. The label of this new image is decided as the same to the label of its nearest neighbour (with the maximum similarity value) in the training set.}


\subsection{Dam reservoir extraction}
The overall pipeline for dam reservoir extraction connects the water body segmentation with dam reservoir recognition via a connected area (CA) extractor: individual water bodies are extracted from the output of the segmentor and fed as input to the water body classifier to predict their labels.
{The CA extractor considers two pixels connected if they are neighbors and have the same value in the predicted segmentation mask. For every pixel, we look into its neighbors in the 3 $\times$ 3 region centered at this pixel.     
Different water body segments are extracted from the image. \miaojing{The smallest water body in our dataset has an area over 200 pixels. Given an image, we set a threshold of 20 pixels to filter out any predicted segment whose area is below 20 pixels. These tiny segments are too small to be water bodies and are therefore considered as noises.
We choose this threshold to be rather small to avoid filtering out real water bodies, as we anticipate in an open-set problem there exist water bodies with areas bigger than 20 pixels but smaller than 200 pixels.}
A bounding box is placed onto each extracted segment according to its pixel coordinate extent. To allow more context for the follow-up classification task, we extend the bounding box by a factor of 2 and crop the corresponding image patch as the input to the classifier.  }

\section{Experiments}
\subsection{Dataset}
\begin{figure}[t]
\centering
\includegraphics[scale=.5]{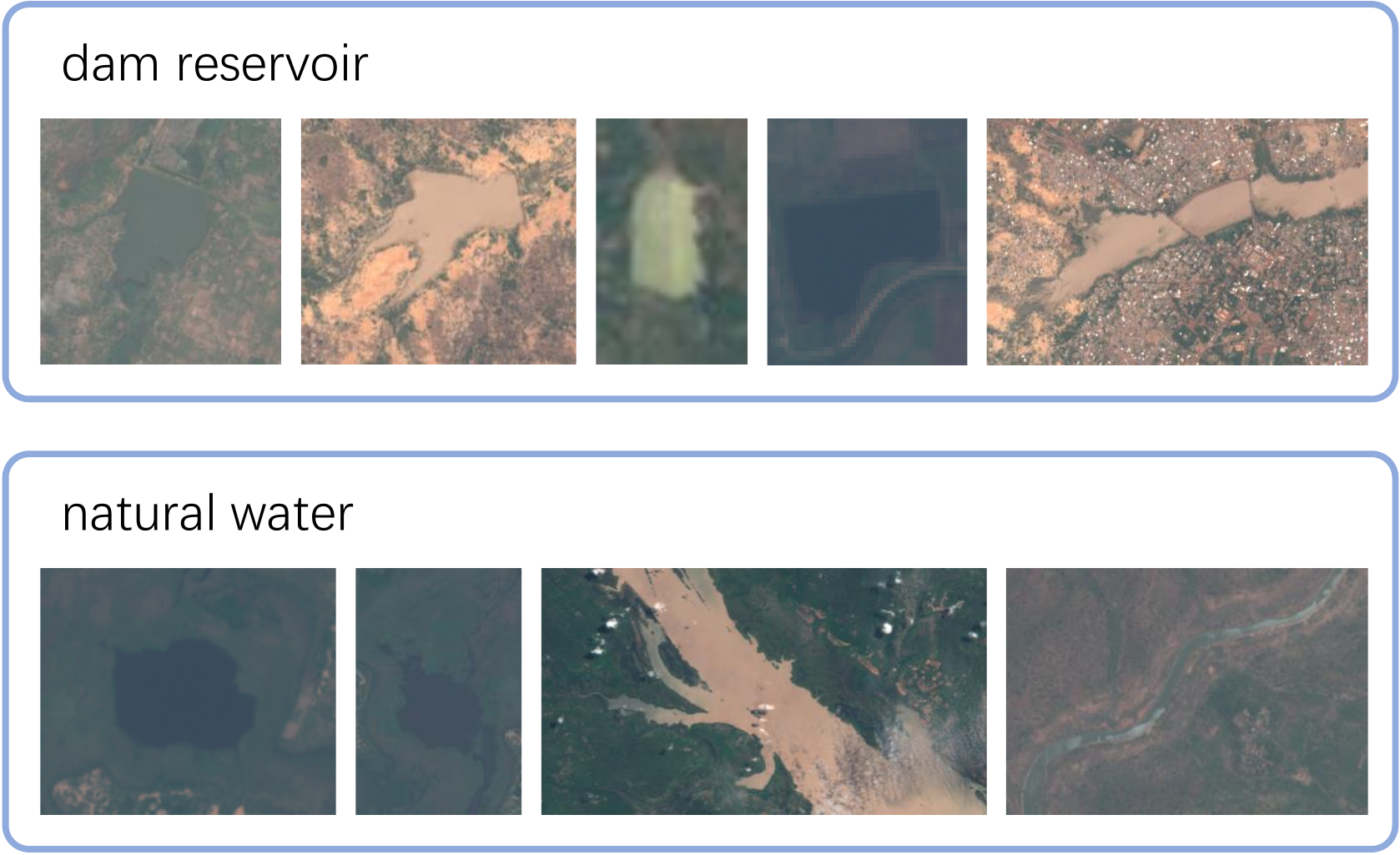}
\caption{Samples for dam reservoir recognition.}
\label{fig:show_cls}
\end{figure}

Having the data samples introduced in Sec.~\ref{sec:watersource}, we curate them into two subsets for water body segmentation and dam reservoir recognition tasks, respectively. In the segmentation set, we keep the original size of the samples cropped in the GIS. Each sample is associated with a ground truth mask whose pixels have values 0 for background lands and 1 for water areas (see Fig.~\ref{fig:data}).
In the classification set, we further crop the training samples in the segmentation set into individual water bodies with bounding boxes (see Fig.~\ref{fig:show_cls}). Instead of having a tight bounding box around each water body, we extend it by a factor of 2 to include some context for classification. We give a label of 0 for the natural water body and 1 for the dam reservoir. For the dam reservoir extraction task, it is performed on the segmentation set yet targeting to both segment the water bodies and classify them as either dam reservoirs or natural water bodies.

\miaojing{Finally, there are in total 295 images containing 670 water bodies in our segmentation set. We use 177 images for training, 59 images for validation and 59 images for testing. The average number of water bodies per image is 2.27.}
The classification set consists of a training set of 424 images (380 dam reservoirs and 44 natural water bodies), a validation set of 150 images (107 dam reservoirs and 43 natural water bodies) and a test set of 96 images (81 dam reservoirs and 15 natural water bodies).


\subsection{Experimental setup}\label{sec:exp_setup}

\para{Implementation details.}
The segmentation model takes DeepLabV3+~\cite{chen2018eccv} as the backbone. We use an input image resolution of $256\times256$, batch size of 4, initial learning rate of 0.0003, and apply the polynomial decay with a factor of 0.9. The network is optimized by the Adam optimizer with 150 epochs in total. Data augmentation is applied to training images, which includes random horizontal and vertical flip, random brightness change, random rotation and random channel shift.
By default, the number of triplets per image $K$ is set to 50, $\beta$ in (\ref{eq:seg_triplet}) is 0.01, $\alpha$ and $\gamma$ in (\ref{eq:seg_loss}) are 0.25 and 2.0, $\sigma$ in (\ref{eq:loss_seg}) is 0.01.

The classification model takes ResNet50V2~\cite{he2016identity} as the backbone, which is pretrained on the ILSVRC classification task. We use an input image resolution of $224\times224$, batch size of 64, and learning rate of 1e-4. The network is optimized by the Adam optimizer with 400 epochs in total. Each image feature is l2-normalized. By default, we set the number of clusters per batch $Z$ as 4, the margin $\epsilon$ in (\ref{eq:cls_triplet}) as 0.01.

\miaojing{Our experiments were carried out on the deep learning platform Tensorflow 2.6.0 using an NVIDIA TITAN RTX GPU. The operating system is Linux - Red Hat 4.8.5-44.}



\begin{figure*}[htb]
\centering
\includegraphics[scale=.625]{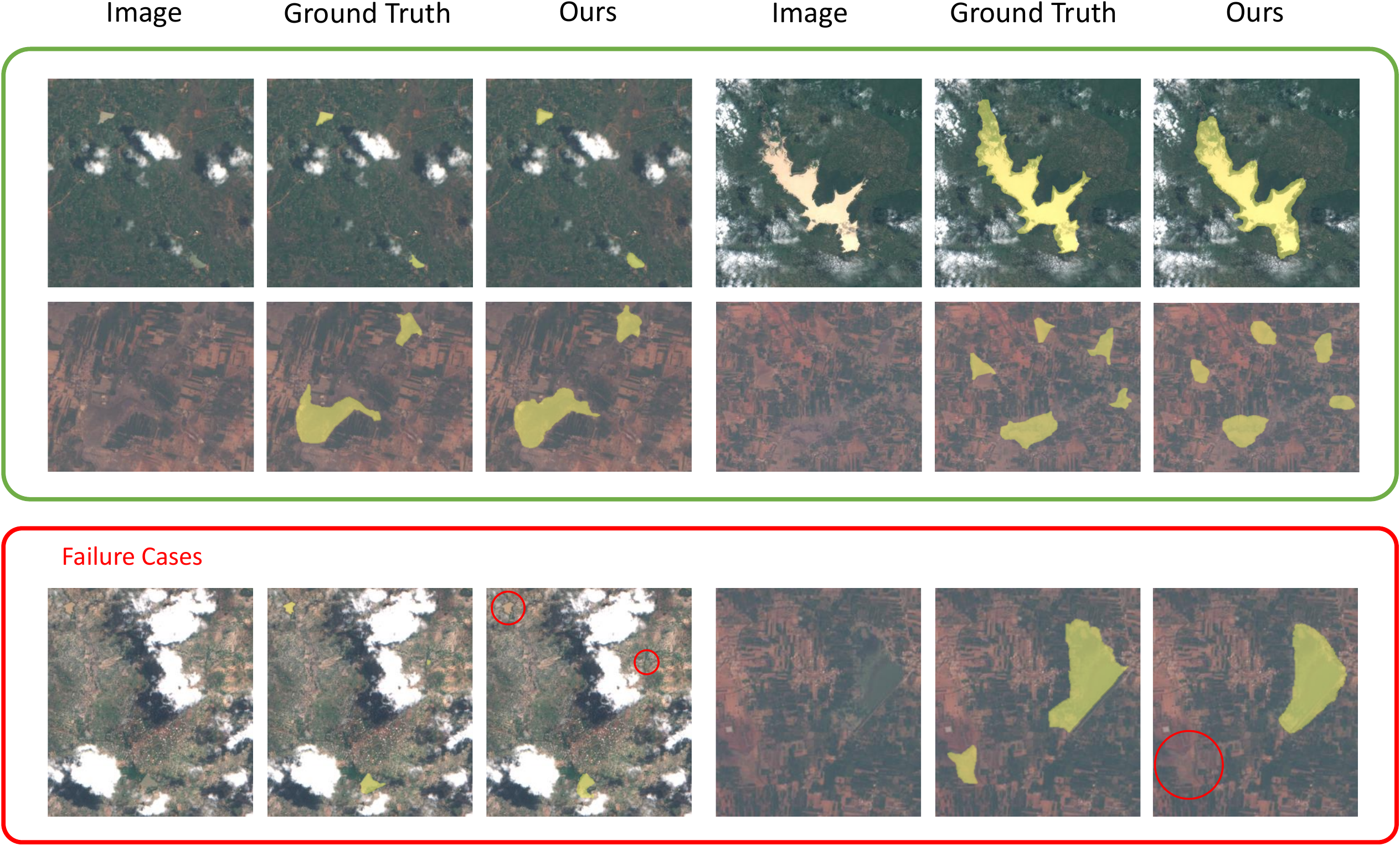}
\caption{Visualization of our water body segmentation results compared with the ground truth. We also provide some failure cases.}
\label{fig:show_seg}
\end{figure*}

\para{Evaluation protocol.}
\miaojing{For dam reservoir recognition, it is a classification task, we compute its accuracy as the ratio of correctly predicted images and all images}. While for water body segmentation and dam reservoir extraction, we evaluate them as a multi-class segmentation problem. For the former, it is of two classes, water body and background land.  For the latter, it is of three classes: after extracting water bodies in the segmentation task, those pixels that are not predicted as water areas are of the class label of background land; as for each extracted water area, if it is classified dam reservoir/natural water body, all pixels within this area are of the same class label. We employ the commonly used Intersection over Union (IoU) metric~\cite{long2015tpami} to evaluate the segmentation result, which computes the ratio of the intersection and union of predicted area and ground truth area for certain class. We take the mean IoU (mIoU) over all classes within each image. This value is further averaged over the entire set.

{For both water body segmentation and dam reservoir extraction, the class of background land is often a high proportion of the image, which can dominate the mIoU value. Hence, in some works~\cite{cheng2017volume}, it is removed from the mIoU.
Furthermore, for dam reservoir extraction, the IoU of dam reservoir and natural water body are averaged with the same weights, while the primary objective of this paper is to identify dam reservoirs. To address these issues, we comprehensively evaluate our experiments: for water body segmentation, we report IoU for only water body by default and mIoU for both water body and background land. For dam reservoir extraction, we report IoU$^\text{d}$ for only dam reservoir (default), mIoU$^\text{dn}$ for dam reservoir and natural water body, mIoU$^\text{dnb}$ for dam reservoir, natural water body, and background land. }

\subsection{Water body segmentation}
\para{Comparison to state of the art.}
We compare our method, point-level metric learning with triplets across images (PLML), to the following segmentation methods:
\begin{compactitem}
 \item DenseASPP~\cite{yang2018denseaspp} connects a set of atrous convolutional layers in a dense way to generate multi-scale features for segmentation.

 \item SegNet~\cite{badrinarayanan2017segnet} consists of an encoder and decoder followed by a pixel-wise classification layer. The pooling indices computed in the max-pooling layer of the encoder are used to perform non-linear upsampling in the decoder.

 \item UNet~\cite{ronneberger2015u}  has a U-shaped architecture that hierarchically connects the contracting part with a symmetric expanding part to incorporate more contextual information.

\item BiSeNet~\cite{yu2018bisenet}  designs a special spatial path to preserve the spatial information for high-resolution features and a context path to perform a fast downsampling upon the former. 

\item PSPNet~\cite{zhao2017pyramid} exploits the capability of global context information by different region-based context aggregation through a pyramid pooling module.

\item Gated-SCNN~\cite{takikawa2019gated} proposes a new two-stream DNN architecture to incorporate  shape information into the network.

\item STDC~\cite{Fan_2021_CVPR} {improves BiSeNet~\cite{yu2018bisenet} by removing its structure redundancy via a new and efficient short-term dense concatenation network.}

\item DeepLabV3+~\cite{chen2018eccv} extends the former DeeplabV3~\cite{chen2017rethinking} by refurbishing its decoder module to refine segmentation results particularly on object boundaries.
\end{compactitem}



DeepLabV3+ is our default backbone. It performs the best compared to the other state of the art methods in Table~\ref{seg-compare}. When we inject our PLML into it, PLML + DeepLabV3+ significantly outperforms all methods. Particularly, it improves DeepLabV3+ by +3.6\% on IoU. PLML is particularly good at improving the segmentation on ambiguous edges between  water and  land areas. We also present the results of mIoU in Table~\ref{seg-compare}. As suggested in evaluation protocol, they are in general higher than those of IoU. The same observation regarding the PLML is found on the mIoU metric. Some qualitative results are presented in Fig.~\ref{fig:show_seg}, our results are very close to the ground truth.

\begin{table}[t]
 \setlength{\tabcolsep}{3.3pt}
 \centering
 \caption{{Comparison to state of the art on water body segmentation.} }
 \label{seg-compare}
 \begin{tabular}{ccc}
  \toprule
  method              & IoU        &{mIoU} \\
 \midrule
    DenseASPP~\cite{yang2018denseaspp}          & 0.364   &{0.623}       \\
    SegNet~\cite{badrinarayanan2017segnet}        & 0.435  &{0.669}        \\
    UNet~\cite{ronneberger2015u}        & 0.446  & {0.677}        \\
    BiSeNet~\cite{yu2018bisenet}               & 0.465  & {0.694}        \\
    PSPNet~\cite{zhao2017pyramid}  & 0.465   &{0.694}       \\
    Gated-SCNN~\cite{takikawa2019gated}  & 0.467  & {0.695}        \\
    STDC~\cite{Fan_2021_CVPR} & 0.468 &{0.699} \\
    DeepLabV3+~\cite{chen2018eccv}           & 0.471  & {0.705}       \\
    PLML+DeepLabV3+(Ours)            & \textbf{0.507} & \textbf{0.742}\\
  \bottomrule
 \end{tabular}
\end{table}

\para{Ablation study.} Our essential contribution is the point-level metric learning with triplets across images. If we remove it from the network, it degrades to the DeepLabV3+ method, with a drop of 3.6\% from 0.507. Next, we study two key elements in PLML: 1) point-level feature position; 2) point-level triplet selection.

\noindent \emph{Point-level feature position.} Features taken from different positions of the network can be utilized as point-level features for the metric learning. We specifically experiment with three positions: following up the design of DeepLabV3+, we take the features by the end of entry flow, middle flow and exit flow, respectively (see \cite{chen2018eccv}); and denote them by PLML (w/ ef), PML (w/ mf) and PLML (w/ exf) in Table~\ref{seg-position}.
Using features by the end of the entry flow gives us a very low IoU of {0.277}. In contrast, using features by the end of the middle flow produces an IoU of 0.474 while features by the end of the exit flow produces the best IoU of 0.507.
The feature by the end of the exit flow contains primarily high-level information and is richer in terms of the channel size compared to the other two positions. 

\begin{table}[!t]
 \setlength{\tabcolsep}{3.3pt}
 \centering
 \caption{{Ablation study on point-level feature position for water body segmentation.} }
 \label{seg-position}
 \begin{tabular}{cccc}
  \toprule
 Variant &   PLML (w/ ef)          &    PLML (w/ mf)   & PLML (w/ exf) \\
     \midrule
  IoU & {0.277} & 0.474 & \textbf{0.507} \\
  \bottomrule
 \end{tabular}
\end{table}

\begin{table}[!t]
 \setlength{\tabcolsep}{3.3pt}
 \centering
 \caption{{Ablation study on point-level triplet selection strategy for water body segmentation.} }
 \label{seg-ablation}

 \begin{tabular}{cccc}
  \toprule
 Variant &   PLML (w/ twi)          &    PLML (w/ fs)   & PLML \\
     \midrule
  IoU & 0.484 & 0.480 & \textbf{0.507} \\
  \bottomrule
 \end{tabular}
\end{table}

\begin{figure*}[htb]
\centering
\includegraphics[scale=.53]{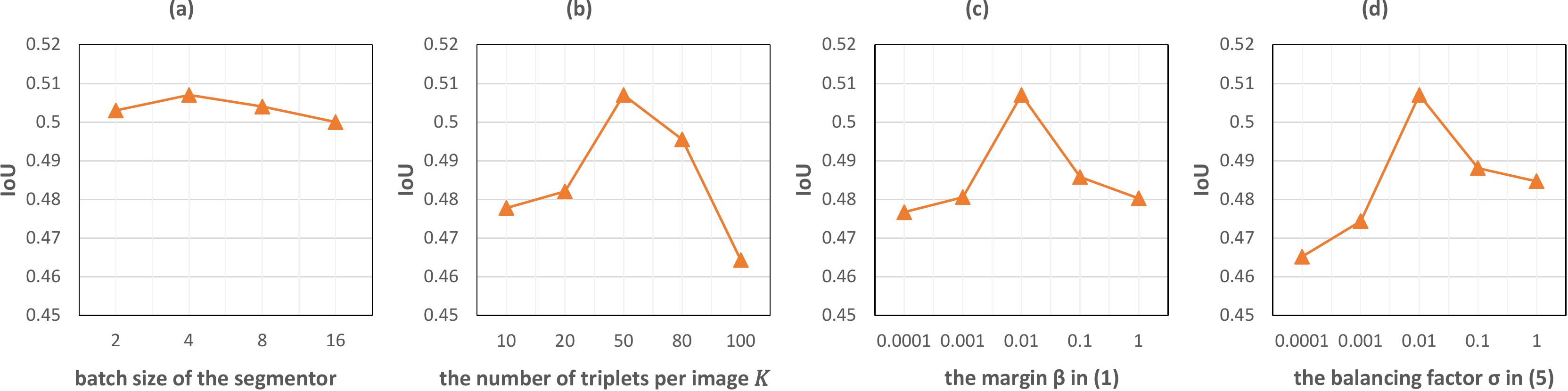}
\caption{Parameter variation on water body segmentation: \miaojing{(a) the batch size of the segmentation model}; (b) the number of triplets per image $K$; (c) the margin in (\ref{eq:seg_triplet}); (d) the loss balancing term $\sigma$ in~\ref{eq:loss_seg}. When analysing one parameter, we fix the other parameters by their default settings. }
\label{parameters_seg}
\end{figure*}

\noindent \emph{Point-level triplet selection. }
Our PLML is realized based on triplets across images. Given an anchor point, its hard positive and hard negative points can be selected across images. To justify this design, we provide the result by confining the triplet construction within each image in Table~\ref{seg-ablation}. We use PLML(w/ twi) to denote the PLML with triplets within images. It is clearly inferior to our original PLML with a decrease of 2.3\% on IoU. As discussed in Sec.~\ref{sec:water_body_segmentation}, PLML with triplets across images can improve the model generalization towards seasonal changes on the appearance of water areas. 

Second, given an anchor point, its hard positive and negative points are randomly selected from false negative and false positive points over the batch, respectively. A classic way is to compute the feature distances from the anchor to all positive and negative points over the batch. The hard positive is of the farthest distance among all positives while the hard negative is of the nearest distance among all negatives. We compare our PLML with the feature-based selection strategy PLML(w/ fs) in Table~\ref{seg-ablation}. PLML(w/ fs) achieves an IoU of 0.48, lower than our PLML. The feature-based strategy is computationally slow, and may also suffer from outliers.


\para{Parameter variation.} We evaluate the parameter variation regarding 1) the batch size of the segmentation model; 2) the number of triplets per image $K$; 3) the margin $\beta$ in (\ref{eq:seg_triplet}); 4) the loss weight $\sigma$ in (\ref{eq:loss_seg}).

\noindent \emph{\miaojing{Batch size of the segmentation model.}} \miaojing{The segmentation model produces pixel-level class predictions which requires a lot of memory. We vary its batch size over 2, 4, 6, and 16  in Fig.~\ref{parameters_seg} (a) where it shows that our default setting of 4 works the best.}

\noindent \emph{Number of triplets per image $K$.} We vary $K$ from 10 to 100 and report the IoU in Fig.\ref{parameters_seg} (b). It can be seen that the IoU increases when $K$ increases from 10 to 50, and decreases afterwards. $K = 50$ achieves the best IoU. Having too many triplets will not only impair the performance but also incur additional cost.

\noindent \emph{$\beta$ in (\ref{eq:seg_triplet}).} We vary $\beta$ over the values 0.0001, 0.001, 0.01, 0.1 and 1 in Fig.\ref{parameters_seg} (c). The best result occurs at $\beta = 0.01$. Referring to (\ref{eq:seg_triplet}), $\beta$ requires $d(a,p)$ to be smaller than $d(a,n) - \beta$. If it is too big, the loss would be too hard to optimize. If it is too small, it might be too easy to learn nothing.

\noindent \emph{$\sigma$ in (\ref{eq:loss_seg}).} We vary $\sigma$ over 0.0001, 0.001, 0.01, 0.1 and 1 in Fig.\ref{parameters_seg} (d) and find the best performance at $\sigma = 0.01$. $\sigma$ balances the loss weight between the focal loss on segmentation and the triplet loss on metric learning. The former is the principal optimization force.

\subsection{Dam reservoir recognition}

\para{Comparison to state of the art.} We compare our proposed method, prior-guided metric learning with triplets from clusters (PGML), to the following  classification methods:

\begin{compactitem}
\item VGG16~\cite{simonyan2014very} increases DNN depth using an architecture with many 3x3 convolution filters stacked together for 16 layers in total. 

\item EfficientNet~\cite{tan2019efficientnet} proposes a new scaling method to better balance network depth, width, and resolution. 

\item InceptionV3~\cite{szegedy2016rethinking}  utilizes batch normalization and factorization into the Inception architecture, which improves the computation efficiency and network accuracy.

\item Inception-ResNet-V2~\cite{szegedy2017inception} integrates the residual learning of ResNet into the Inception net.

\item Xception~\cite{chollet2017xception} is an extension of the Inception net, which replaces the standard Inception block with depthwise separable convolutions.

\item ResNet50V2~\cite{he2016identity} utilizes identity mapping to improve the original ResNet~\cite{he2016deep} (\ie ResNet34 and ResNet50) and achieves a better performance.

\item EfficientNetV2~\cite{tan2021efficientnetv2}: { uses neural architecture search to jointly optimize the model size and training speed. }

\end{compactitem}

ResNet50V2~\cite{he2016identity} is our default backbone, it  produces the second best result among other comparable methods in Table \ref{class-compare}. \miaojing{Particularly, ResNet50 and ResNet34 were utilized in previous dam reservoir recognition studies~\cite{fang2019recognizing,malerba2021continental}.} When we inject our PGML into ResNet50V2, it gives us +4.17\% improvement and reaches the highest accuracy 0.99. We also review a very recent work, EfficientNetV2~\cite{tan2021efficientnetv2},  which performs the best over previous state of the art. When we inject the PGML into EfficientNetV2, it further improves +1.10\%. This is lower than PGML+ResNet50V2, but still shows the effectiveness of PGML. We list the number of learnable parameters in different networks. PGML brings no additional parameters to ResNet50V2/EfficientNetV2. 

\begin{table}[!t]
 \setlength{\tabcolsep}{3.3pt}
 \centering
 \caption{{Comparison to state of the art on dam reservoir recognition.} }
 \label{class-compare}

 \begin{tabular}{ccc}
  \toprule
  method              & accuracy    & params    \\
 \midrule
    VGG16~\cite{simonyan2014very}               & 0.927      &   134.30M \\
    EfficientNet~\cite{tan2019efficientnet}        & 0.927      &    4.05M  \\
    InceptionV3~\cite{szegedy2016rethinking}            & 0.938      &   21.80M \\
    Inception-ResNet-v2~\cite{szegedy2017inception} & 0.938      &   54.60M \\
    Xception~\cite{chollet2017xception}            & 0.938      &    20.90M\\
    ResNet34~\cite{malerba2021continental} & 0.927 & 22.36M \\
    ResNet50~\cite{fang2019recognizing} & 0.938 & 23.59M \\
    ResNet50V2~\cite{he2016identity}          & 0.948      &    23.56M\\
    EfficientNetV2~\cite{tan2021efficientnetv2}          & 0.968      &    21.50M\\
   PGML+EfficientNetV2(Ours)    & 0.979      &    21.50M\\
    PGML+ResNet50V2(Ours)                & \textbf{0.990}    &    23.56M\\
  \bottomrule
 \end{tabular}
 \vspace{-3mm}
\end{table}

\para{Ablation study.} Our key contribution is the prior-guided metric learning with triplet from clusters. To validate its effectiveness, we first compare it with a conventional classification network optimized using the cross entropy loss. This is indeed the accuracy for the ResNet50V2 in Table~\ref{class-ablation}. PGML is clearly better than ResNet50V2, showing that metric learning is more suitable to address variations of water bodies.

Regarding the metric learning, we have specifically constructed triplets via a prior-guided clustering scheme. We compare this with the classic feature-based metric learning scheme (FBML)~\cite{schroff2015facenet}. Given an anchor image, its hard positive is selected as the one containing the same class and having the biggest distance to it; its hard negative is the one containing different classes while having the smallest distance to it. FBML produces a result of 0.958, which is better than using the CE yet clearly lower than our PGML.


\begin{table}[!t]
 \setlength{\tabcolsep}{3.3pt}
 \centering
 \caption{Ablation study on dam reservoir recognition. CE denotes for the cross entropy loss, while FBML feature-based metric learning. }
 \label{class-ablation}
  \begin{tabular}{cccc}
  \toprule
 Variant &   CE  &   FBML   & PGML \\
     \midrule
  Acc & 0.948 & 0.958 & \textbf{0.990} \\
  \bottomrule
 \end{tabular}
\end{table}


\begin{figure}[t]
\centering
\includegraphics[scale=.53]{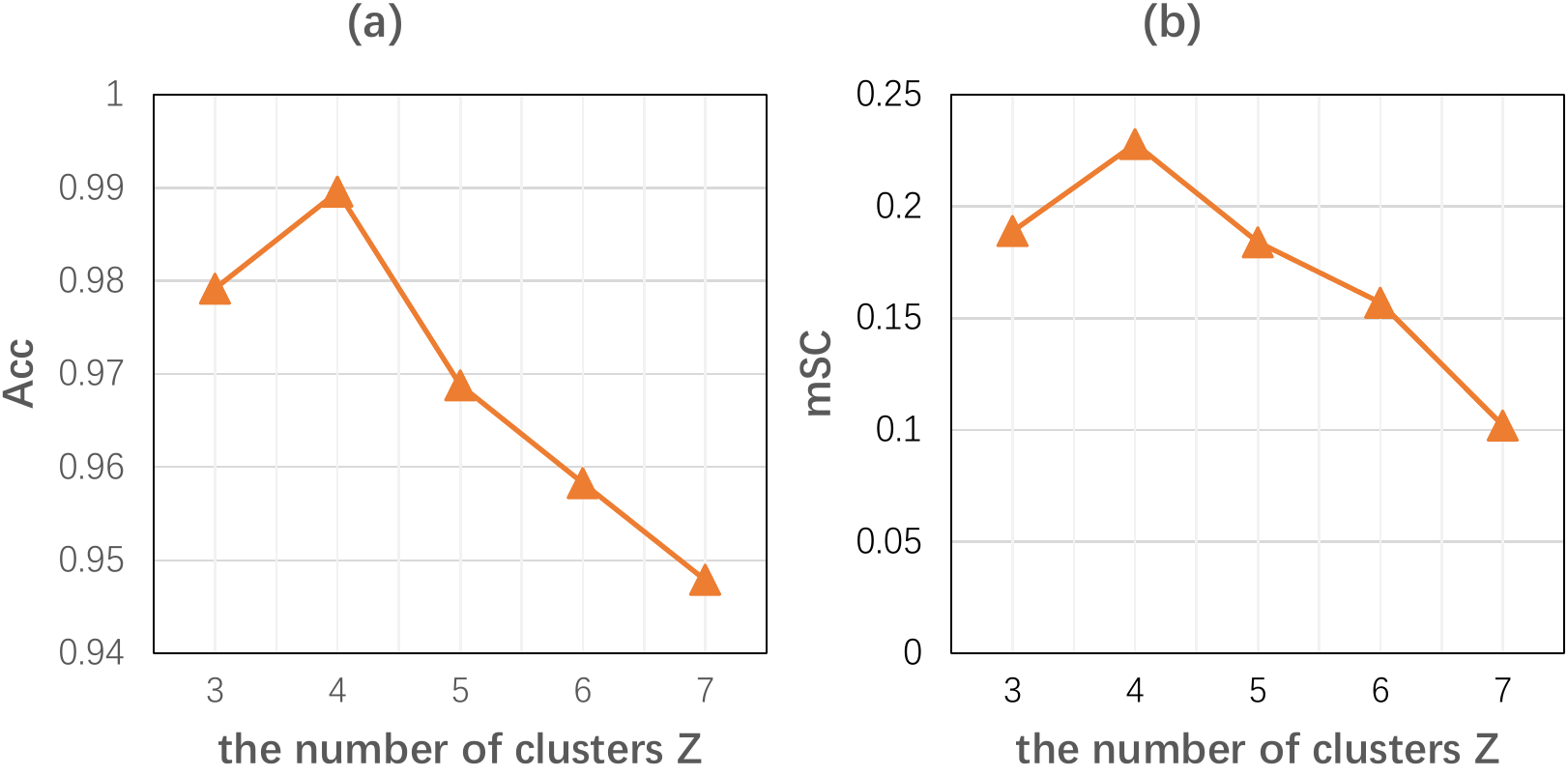}
\caption{The parameter variation on the number of clusters $Z$: (a) using the $Acc$ metric; (b) using the $mSC$ metric. }
\label{parameters_clusters}
\end{figure}

\begin{figure*}[t]
\centering
\includegraphics[scale=.53]{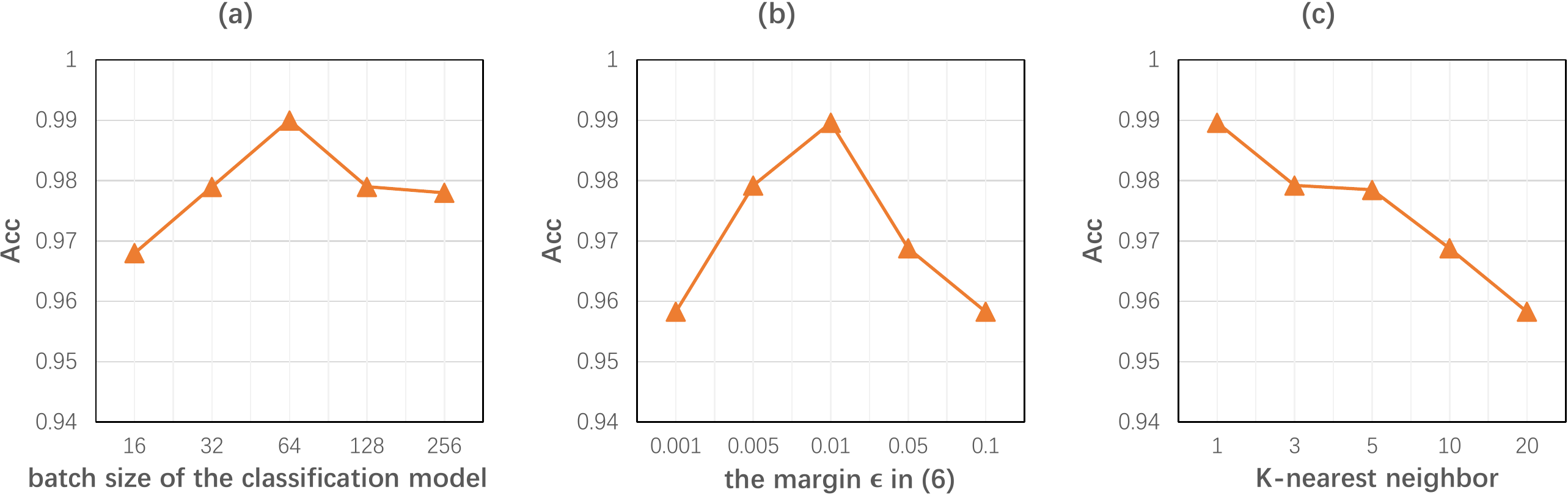}
\caption{Parameter variation on dam reservoir recognition: (a) the batch size of the classification model; (b) the margin $\epsilon$ in (\ref{eq:cls_triplet}); (c) $K$-nearest neighbor at inference. When analysing one parameter, we fix the other parameters by their default settings. }
\label{parameters_class}
\end{figure*}

\para{Parameter variation.} For the classification model, we vary 1) the bath size of the classification model; 2) the number of clusters per batch $Z$, 3) the margin $\epsilon$ in (\ref{eq:cls_triplet}), and 4) $K$-nearest neighbors during inference.

\noindent \miaojing{\emph{Batch size of the classification model.} The classification model produces image-level class predictions which requires not much memory. We vary its batch size over 16, 32, 64, 128, and 256 in Fig.~\ref{parameters_class} (a) where it shows that our default setting of 64 works the best. }

\noindent \emph{Number of clusters $Z$.} Inspired by the prior information (\eg shapes) of water bodies, we know $Z$ should not be very big particularly within one batch of images. We vary it between 3 and 7 and report the results in Fig.\ref{parameters_clusters} (a). The best performance is at $Z = 4$, which is our default setting.

\miaojing{Apart from using the classification accuracy to choose the optimal $Z$, we also adopt the Silhouette Coefficient~\cite{rousseeuw1987silhouettes} to evaluate the effectiveness of the clustering. The Silhouette Coefficient is a widely used metric to measure the quality of clustering methods~\cite{hasanzadeh2018learning,matioli2018new}. It is composed of two scores: given a sample $x$, $a$ is the mean distance between $x$ and all other samples in the same cluster to $x$; $b$ is the mean distance between $x$ and all other samples in the next nearest cluster to $x$. We compute the Euclidean distance between the features of any two samples. The Silhouette Coefficient $SC$ for a single sample is defined as:
\begin{equation}\label{eq:sc}
SC = \frac{b-a}{max(a,b)}
\end{equation}
The Silhouette Coefficient for a batch of samples is computed as the average of $SC$ over the batch. This score is bounded between -1 for incorrect clustering and +1 for highly dense clustering. In general, when $SC$ is positive, we can assume the resulting clusters are well separated~\cite{moura2014analysis}. We compute the mean average of $SC$ over batches and denote it as $mSC$ in Fig.~\ref{parameters_clusters} (b). By varying the number of clusters $Z$, $mSC$ reaches the maximum value of 0.228 when $Z = 4$. This is indeed consistent with the best $Z$ shown in  Fig.~\ref{parameters_clusters} (a). }

The margin $\epsilon$ (\ref{eq:cls_triplet}) used in the triplet loss is set to 0.01 by default. The experiment shown in Fig.\ref{parameters_class} (b) validates this. Similar to $\beta$ in (\ref{eq:seg_triplet}),   $\epsilon$
can be neither too small nor too big.

\noindent \emph{$K$-nearest neighbor.} As mentioned in Sec.~\ref{sec:dam-recog}, once the classification model is trained, the test image label is inferred through its nearest neighbor in the training set. If we instead use the $K$-nearest neighbors with $K$ increased from 1, 3, 5, 10, and 20, we observe in Fig.~\ref{parameters_class} (c) that the performance gets decreased. $K = 1$ works the best in practice.

\subsection{Dam reservoir extraction}\label{sec:exp-overall}
Referring to Sec.~\ref{sec:exp_setup} the evaluation protocol, we report IoU$^\text{d}$, mIoU$^\text{dn}$, and mIoU$^\text{dnb}$ for the overall dam reservoir extraction pipeline in Table~\ref{whole-compare}. mIoU$^\text{dn}$ measures the segmentation accuracy equivalently on dam reservoir and natural water body, which produces a value of 0.453. It is only slightly lower than the IoU value of PLML+DeepLabV3+ in Table~\ref{seg-compare}, which shows the robustness of the water body classifier based on imperfect water body segments. Since the extraction of dam reservoir is more important than natural water body, if we only look at the result of IoU$^\text{d}$, it is actually higher than mIoU$^\text{dn}$ (0.538 \vs 0.453). This shows a particular strength of our model predicting on dam reservoirs. It is reasonable as dam reservoirs generally share more common traits than natural water bodies.

The ablation study is offered in the table to validate the effectiveness of the proposed PLML and PGML. For instance, without using PLML and PGML, the pipeline degrades to the combination of DeepLabV3+ and ResNet50V2, which produces an IoU$^\text{d}$ of 0.434,  mIoU$^\text{dn}$ of 0.331, and mIoU$^\text{dnb}$ of 0.53, 10.4\%, 12.2\%, and 8.5\% lower than the full version of ours (\ie 0.538, 0.453, and 0.615), respectively.

\miaojing{Our pipeline solves the problem in two stages, where we take bounding boxes of predicted water body segments from the first stage to classify them as either natural water bodies and dam reservoirs in the second stage.  To justify this design, we offer two comparisons. First, instead of using bounding boxes of segments, we feed the segments of water bodies directly into the classification model. We denote this variant as Our pipeline (Seg) in Table~\ref{whole-compare}, where it shows that Our pipeline (Seg) performs clearly inferior to the original pipeline. We argue that having certain extent background information included into the bounding box of a segment indeed provides context for the water body, which can help the water body classification. Next, we compare our pipeline with a one-stage end-to-end multi-class semantic segmentation network. We employ the multi-class DeepLabV3+ network (MC-DeepLabV3+) to directly predict pixel-level labels, \ie dam reservoir, natural water body, and background land. Its performance in Table~\ref{whole-compare} is significantly lower than our two-stage pipeline. As argued in Sec.~\ref{Sec:intro}, a multi-class segmentation network can not work well as there exists not much pixel-level difference of water areas between dam reservoirs and natural water bodies. }

\miaojing{Despite the good results of our method, there still exist several factors in general that limit the performance of dam reservoir extraction. For instance, the water body size varies drastically among different samples. Clouds occur in some samples which occlude water bodies. The contour ambiguity between the water area and land region is also an impediment. Moreover, there may exist a distribution shift between samples in the training set and test set. We will keep investigating these factors in our future work. }

\begin{figure}[t]
\centering
\includegraphics[scale=.48]{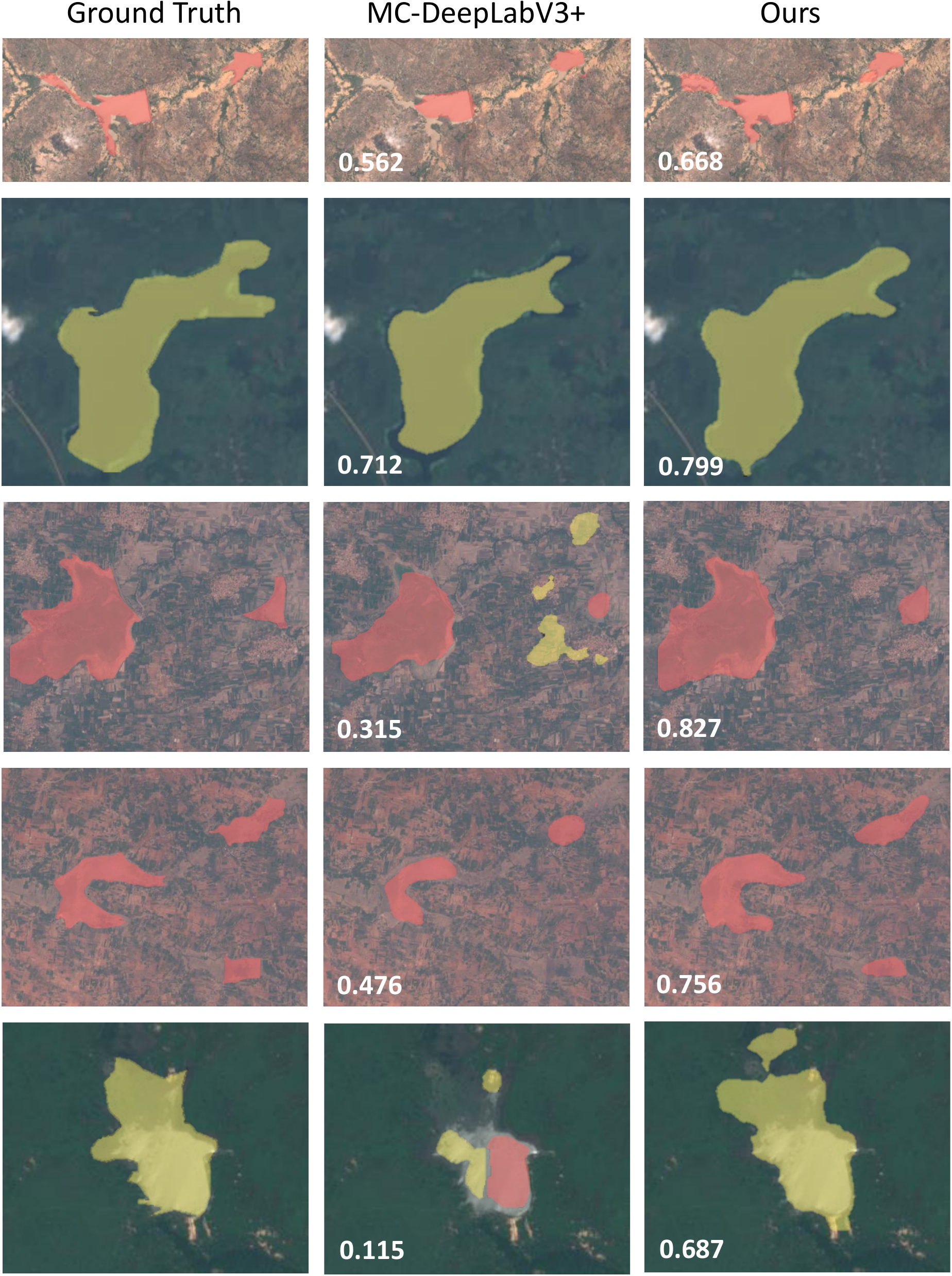}
\caption{{Visualization of dam reservoir extraction results using our pipeline and MC-DeepLabV3+. Ground truth is also provided. Red pixels denote the dam reservoir class while {yellow} ones denote the natural water body class. \miaojing{The value of mIoU$^\text{dn}$ is given in each image.} }}
\label{fig:final_seg}
\end{figure}

Some qualitative results are shown in Fig.~\ref{fig:final_seg} for our pipeline and MC-DeepLabV3+. Compared to the ground truth, ours is clearly better than the MC-DeepLabV3+.

\begin{table}[t]
 \setlength{\tabcolsep}{3.3pt}
 \centering
 \caption{{Performance on overall dam reservoir extraction.} }
 \label{whole-compare}
 \begin{tabular}{cccccc}
  \toprule
  method              & PLML    & PGML & IoU$^\text{d}$ & mIoU$^\text{dn}$ & mIoU$^\text{dnb}$\\
 \midrule
  \multirow{4}{*}{Our pipeline} &  \xmark  &  \xmark& 0.434 &{0.331} &  {0.530} \\
   & \cmark   &\xmark & 0.457 & {0.366} &  {0.556} \\
 & \xmark   & \cmark  & 0.470& {0.394} & {0.572} \\
    & \cmark    & \cmark & \textbf{0.538} & \textbf{0.453} & \textbf{0.615} \\
    \midrule
    Our pipeline (Seg)  & \cmark    & \cmark & 0.497 & 0.409 & 0.587\\
        MC-DeepLabV3+ & n/a & n/a & 0.408&  {0.305} & {0.498}  \\
  \bottomrule
 \end{tabular}
\end{table}

\section{Conclusion}
We developed a novel deep learning pipeline to segment water bodies in high resolution remote sensing imagery and subsequently classify them as either dam reservoir or natural water body. For water body segmentation, we introduce a point-level metric learning (PLML) scheme with triplets across images. 
For dam reservoir recognition,  we introduce a prior-guided metric learning scheme with triplets from clusters.
A connected area extractor is devised onto the output of the water body segmentor to feed individual water bodies into the water body classifier. The whole pipeline allows for the
separation of natural water bodies from 
dam reservoirs in publicly available global remote sensing data (\ie Sentinel-2 data), even though these data are often too coarse to be used to find small dam reservoirs in previous studies.
To facilitate further research in this field, we have created a benchmark dataset for two geographically distinct regions, the Volta basin in western Africa and the Cauvery basin in India.
Extensive experiments on this dataset demonstrate the effectiveness of our method over state of the art methods.
This work has the potential to be scaled up across other regions around the world to identify millions of dam reservoirs important for smallholder irrigation, support more research on the impacts of drought under climate change, and help advise water management policy.

\section*{Acknowledgment}
This work was supported by a King's Together award, ``Remote Sensing and Machine Learning to Detect Global Dams and Associated Reservoirs", from King's College London; and funding from the European Union’s Horizon 2020 FET Proactive Programme under grant agreement No 101017857 for the project ``Restarting the Economy in Support of Environment, through Technology (ReSET)". Miaojing Shi was partially supported by the National Natural Science Foundation of China (NSFC) under Grant No. 61828602.

\ifCLASSOPTIONcaptionsoff
  \newpage
\fi



\bibliographystyle{IEEEtran}
\bibliography{bibliography.bib}
\end{document}

%% file: macro.tex
\def \ie {{i.e.}~}
\def \eg {{e.g.}~}
\def \etc {{etc.}~}
\def \etal {{et al.}~}
\def \vs {{v.s.}~}

\newcommand{\para}[1]{\noindent \textbf{#1}}